\documentclass[runningheads]{article}

\usepackage{arxiv}

\usepackage[utf8]{inputenc} 
\usepackage[ruled,vlined,linesnumbered,algoruled]{algorithm2e}
\usepackage{algorithmic}
\usepackage[T1]{fontenc}    
\usepackage{hyperref}       
\usepackage{url}            
\usepackage{booktabs}       
\usepackage{amsfonts}       
\usepackage{nicefrac}       
\usepackage{microtype}      
\usepackage{lipsum}
\usepackage{graphicx}
\usepackage[utf8]{inputenc}
\usepackage{caption}
\usepackage{float}
\usepackage{refstyle}
\usepackage[export]{adjustbox}
\usepackage{amsmath}
\usepackage{ulem}
\usepackage{fix-cm}
\usepackage{array}
\usepackage{caption}
\usepackage{subcaption}
\usepackage{longtable}
\usepackage{changepage}
\usepackage{textcomp}
\usepackage{multirow}
\usepackage{fancyhdr}
\usepackage{comment}

\newcommand*\samethanks[1][\value{footnote}]{\footnotemark[#1]}
\begin{document}

\title{Bayesian Optimization - Multi-Armed Bandit Problem}


\author{
 Abhilash Nandy \thanks{Everyone contributed equally.}\\
  Indian Institute of Technology Kharagpur\\
  West Bengal\\
  India \\
  \texttt{nandyabhilash@kgpian.iitkgp.ac.in} \\
  \And
 Chandan Kumar \samethanks\\
  Indian Institute of Technology Kharagpur\\
  West Bengal\\
  India \\
  \And
 Deepak Mewada \samethanks\\
  Indian Institute of Technology Kharagpur\\
  West Bengal\\
  India \\
  \And
 Soumya Sharma \samethanks\\
  Indian Institute of Technology Kharagpur\\
  West Bengal\\
  India \\
}
\maketitle
\begin{abstract}
  In this report we survey Bayesian Optimization methods focussed on Multi-Armed Bandit Problem. We take the help of the paper "Portfolio Allocation for Bayesian Optimization" ~\cite{SirPaper}. We report a small literature survey on the acquisition functions and the types of portfolio strategies used in the papers ~\cite{SirPaper} ~\cite{srinivas2010gaussian}. We also replicate the experiments
  \footnote{Code link: \url{https://colab.research.google.com/drive/1GZ14klEDoe3dcBeZKo5l8qqrKf_GmBDn?usp=sharing##scrollTo=XgIBau3O45_V}} in ~\cite{SirPaper} and report our findings and compare them to the results in the paper. 
\end{abstract}


\section{Introduction} \label{introduction}

Many optimization problems in machine learning are black box optimization problems where the objective function $f(x)$ is a black box function. We do not have an analytical expression for $f$ nor do we know its derivatives. Evaluation of the function is restricted to sampling at a point $x$ and getting a possibly noisy response.

If $f$ is cheap to evaluate we could sample at many points e.g. via grid search, random search or numeric gradient estimation. However, if function evaluation is expensive e.g. tuning hyperparameters of a deep neural network, probe drilling for oil at given geographic coordinates or evaluating the effectiveness of a drug candidate taken from a chemical search space then it is important to minimize the number of samples drawn from the black box function $f$.

This is the domain where Bayesian optimization techniques are most useful. They attempt to find the global optimum in a minimum number of steps. Bayesian optimization incorporates prior belief about $f$ and updates the prior with samples drawn from $f$ to get a posterior that better approximates $f$. The model used for approximating the objective function is called surrogate model. Bayesian optimization also uses an acquisition function that directs sampling to areas where an improvement over the current best observation is likely.

In this report, we try to create an understanding around Bayesian Optimization with a focus on Multi-Arm Bandit Problem. In Section \ref{Literature Survey} we talk about the origins of Bayesian Optimization and the use of the method in different literature, we also talk about different acquisition functions and give a brief overview on the requirement for Multi-Armed Bandit Framework. The literature survey can be considered as a small summary of the paper. In Section \ref{Problem Definition} we define Bayesian Optimization and its characteristics, give an overview of Gaussian Processes and give a brief introduction of different Acquisition functions. In Section \ref{Methodology} we talk about different Portfolio Strategies and give the algorithm for Hedge function. In Section \ref{Experimental Results} we talk about the experimental settings and specify our replication of the results and the experiments in the paper. In Section \ref{Discussions} we talk about the behaviour of convergence and conclude the report in Section \ref{Conclusion}.

\section{Literature Survey}
\label{Literature Survey}
The term "Bayesian Optimization" originated in the 1970s \cite{seventies}, where the objective was to find a solution for the extremum of a function such that the standard deviation w.r.t the extremum is as minimum as possible. This notion was extended to accommodate non-linear function optimization, especially in the field of engineering, by using the technique of exploration-exploitation trade-off of the sample space in order to approximate the function surface using Efficient Global Optimization (EGO) \cite{EGO1}. Since then, Bayesian Optimization has been used in EGO for optimization of black box functions \cite{EGO1}, Gaussian Processes \cite{EGO-GP}, MCMC \cite{mcmc} etc. Bayesian Optimization Approaches are especially helpful when the number of samples that can be sampled from black-box functions are less, due to the process of sampling being either expensive or time-taking. Theoretical Consistency of Bayesian Optimization has been shown in~\cite{consistency1},~\cite{consistency2}. 

Choosing an acquisition function is of primal importance and plays a pivotal role in the rate of convergence of the Optimization procedure. One of the earliest successful acquisition functions is the Probability of Improvement (PI)~\cite{PI}. As the name suggests, function tries sampling such a point in the near future that increases the probability of getting a sample having a  higher output value of the black-box objective function. This evolved into a function, which tries to take into account the increase in the function value as well, thus giving rise to an improved Expected Improvement (EI) Acquisition Function \cite{consistency2}. Gaussian Process Upper-Confidence Bound (GP-UCB) is a relatively newer function, which runs a greedy selection of samples such that the function values have a reasonable upper bound on the actual extremum, thus having an upper hand on the previously mentioned algorithms in terms of the exploration-exploitation trade-off, as it is able to decrease uncertainty of sampling points near extremum and not globally, thus ensuring a better convergence. Some other recent acquisition functions use other techniques like Knowledge Gradient \cite{KG} that allows us to sample points that creates the possibility of a solution that has not been evaluated in the past, Entropy Search \cite{entropysearch} that tries to minimize the decrease in the differential entropy \cite{diffentropy} to reach the extremum, etc. 

However, using individual acquisition functions has shown to be non-conclusive, since the choice of an acquisition function is quite subjective. Hence, portfolio allocation strategies that help in selection of an appropriate ensemble of acquisition functions have been applied, which have shown better results, especially in terms of the gap metric. \cite{SirPaper} proposed a solution in this regard by using the concept of hierarchical multi-armed bandit strategy \cite{multiarmedbandit}, where two levels of multi-armed bandit strategy are applied - one for sampling the acquisition function, and the other for sampling the next point. Hedge Algorithms \cite{portfolio} use a probabilistic action-based reward framework which sits inside the multi-armed bandit framework. \cite{normalhedge} implements NormalHedge, which is a single-line search followed by an update in the intrinsic parameters. \cite{multiarmedbandit} proposes "Exp3" - A Hedge Algorithm that applies to the partial information setting, by reducing the amount of exploration and exploiting the reward only for a particular action. \cite{SirPaper} introduces an extension of this algorithm, known as GP-Hedge, which is based on a full information setting, which evaluates the reward for every possible action in every iteration.

\section{Problem Definition}
\label{Problem Definition}
In this section, we first define the Bayesian Optimization Problem and the requirement for it. We next define the steps required for a Bayesian Optimization i.e. Gaussian Process and Acquisition Functions.

\subsection{Bayesian Optimization}
\label{Bayesian Optimization}
Optimization is an important problem specially in machine learning where we often need to estimate the minimum of an associated cost function. Most often, these functions are easy to estimate such as in the case of logistic regression where the cost function is single-minima convex function and has closed-form derivatives. However, other optimization problems are more challenging and are expensive to evaluate, often taking several minutes or hours. 
One example of this is hyper-parameter searching in neural networks. Optimizing black box objective functions may have challenging characteristics including \textbf{a)} high cost of evaluation, \textbf{b)} multiple local optima, \textbf{c)} no derivatives, \textbf{d)} mixture of multiple discrete and continuous variables often conditional and noisy.   


Bayesian Optimization is a framework which attempts to estimate global optimum in a minimum number of steps. By building a model of the entire function to be optimized, we attempt to include the current estimation and the uncertainty of the that estimation to get an optimum solution. In more formal terms, Bayesian Optimization addresses the problem of finding the parameters $\hat{x}$ that maximize a function $f(x)$ over some domain $\chi$ consisting of finite lower and upper bounds on every variable: 

\begin{equation}
    \hat{x} = argmax_{x \in \chi} [f(x)]
\end{equation}

At iteration $t$, the algorithm learns by choosing parameter $x_t$ and receiving the corresponding function value $f[x_t]$. Goal of the Bayesian Optimization is to build a probabilistic function of the underlying function and find the maximum point on the function using the minimum number of function evaluations. Bayesian Optimization is a sequential search that incorporates both exploration and exploitation and has two important characteristics: 
\begin{itemize}
    \item \textbf{Probabilistic Model of the function}: Bayesian Optimization starts with a prior(initial probability distribution) over the function $f[\bullet]$ to be optimized. This indicates the uncertainty of the function. With each observation of the function $(x_t,f[x_t])$, we learn more and compute the posterior (the distribution over possible functions) over the unknown objective function.
    \item \textbf{An acquisition function}: Computed from the posterior distribution, an acquisition function indicates the desirability of sampling the next point and automatically trades off between exploration and exploitation.
\end{itemize}

The model used for approximating the objective function is called surrogate model. A popular surrogate model for Bayesian optimization are Gaussian processes (GPs). We now attempt to explain the two steps of Bayesian Optimization in brief, namely Probabilistic Modelling of the function through Gaussian Process and Acquisition Functions.

\subsection{Gaussian Process}
\label{Gaussian Process}

A Gaussian Process is a stochastic process where any point \begin{math}\textbf{x} \in R^d \end{math} is assigned a random variable f(x) and where the joint distribution of a finite number of these variables $\rho(f(x_1),…,f(x_N))$ is itself Gaussian.

\begin{equation}
    f(x)\sim GP(m(x),k(x_i,x_j))
\end{equation}

For convenience and without loss of generality, ~\cite{SirPaper} assumes the prior mean to be the zero function $m[x] = 0$ and the variance to be the squared exponential kernel with a vector of automatic relevance determination (ARD) hyperparameters $\theta$:

\begin{equation}
    m[0] = 0
\end{equation}

\begin{equation}
    k(x_i,x_j) = exp({ - } \frac{1}{2}(x_i-x_j)^T diag(\theta)^{ - 2} (x_i-x_j))
\end{equation}

where $dig(\theta)$ is the diagonal matrix with entries $\theta$ along the diagonal. Here we can see that the covariance decreases as a function of distance between the two points.


Given observations \textbf{F}$ = [f[x_1], f[x_2], ... , f[x_t]]$ at $t$ points, we try to make a prediction of a new point $x^*$. Now, by properties of GPs, \textbf{F} and $f^* = f[x^*]$ are jointly Gaussian:

\begin{equation}
\label{GP}
    Pr({\Big[}\frac{f}{f^*}{\Big]}) = Norm{\Big[}0, {\Big[}\begin{array}{cc}
        K & k \\
        k^T & k(x_*, x_*)
    \end{array}{\Big]}{\Big]}
\end{equation}

where $K$ is a $t \times t$ matrix where element $(i,j)$ is given by $k[x_i,x_j]$, k$ = [k(x_t+1,x_1),k(x_t+1,x_2),...,k(x_t+1,x_t)]$.


Since the function values in equation \ref{GP} are jointly normal, the conditional probability $Pr({\Big[}\frac{f}{f^*}{\Big]})$ must also be normal. So, using the standard formula for mean and variancef this conditional distribution:

\begin{equation}
    Pr(\frac{f}{f^*}) = Norm({\mu[x^*], \sigma^2[x^*]})
\end{equation}

where,

\begin{equation}
    \mu[x^*] = k[x^{*},x] {k[x,x]}^{-1} f
\end{equation}
\begin{equation*}
    \sigma^2[x^{*}] = k[x^{*},x^{*}] - k[x^{*},x] k[x,x]^{ - 1} k[x,x^{*}].    
    \end{equation*}

Using this formula, we can estimate the distribution of the function at any new point $x^*$. The best estimate of the function value is given by the mean $\mu[x]$, and the uncertainty is given by the variance $\sigma^2[x]$

\subsection{Acquisition Functions}
\label{Acquisition Functions}

Acquisition functions serve as a guide for the search of the optimum. The acquisition function takes the mean and variance at each point x on the function and computes a value that indicates how desirable it is to sample next at this position. Typically, high values of the acquisition function correspond to high values of the objective function. Now we explain the multiple acquisition functions used in ~\cite{SirPaper}.

The core ideas associated with acquisition functions: i) they are heuristics for evaluating the utility of a point; ii) they are a function of the surrogate posterior; iii) they combine exploration and exploitation; and iv) they are inexpensive to evaluate.

First three acquisition functions described are Probability of Improvement, Expected Improvement, Upper Confidence Bound from ~\cite{SirPaper}. These are most famous and widely used. We also mention a few alternative acquisition functions as found in ~\cite{chapelle2011empirical}, ~\cite{frazier2009knowledge}.

\subsubsection{\textbf{Probability Of Improvement (PI)}}

PI acquisition function chooses the next query point as the one which has the highest probability of improvement over the current max $f(x^+)$. 

Mathematically, we write the selection of next point as follows,

\begin{equation}
    x_{t+1} = argmax(\alpha_{PI}(x))
            = argmax(P(f(x) \geq (f(x^{+}) + \epsilon))
\end{equation}

where,
$P(\cdot)$  indicates  probability

$\epsilon$   a trade-off parameter, $\epsilon\geq0$

And, $x^+ = \text{argmax}_{x_i \in x_{1:t}}f(x_i)$  where $x_i$
is the location queried at $i^{th}$  time step.

Looking closely, we are just finding the upper-tail probability (or the CDF) of the surrogate posterior. Moreover, if we are using a GP as a surrogate the expression above converts to,

$x_{t+1} = argmax_x \Phi\left(\frac{\mu_t(x) - f(x^+) - \epsilon}{\sigma_t(x)}\right)$
where,

$\Phi(\cdot)$ indicates the CDF

PI uses $\epsilon$ to strike a balance between exploration and exploitation. Increasing $\epsilon$ results in querying locations with a larger $\sigma$ as their probability density is spread.

\subsubsection{\textbf{Expected Improvement (EI)}}
Probability of improvement only looked at how likely is an improvement, but, did not consider how much we can improve. The next criterion, called Expected Improvement, does exactly that! The idea is fairly simple - choose the next query point as the one which has the highest expected improvement over the current $max f(x^+)$, where $x^+ = argmax_{x_i \in x_{1:t}}f(x_i)$ and $x_i$ is the location queried at $i^{th}$ time step.

In this acquisition function, $t + 1^{th}$ query point, $x_{t+1}$ , is selected according to the following equation.

\begin{equation}
x_{t+1} = argmin_x \mathbb{E} \left( ||h_{t+1}(x) - f(x^\star) || \ | \ \mathcal{D}_t \right)
\end{equation}

Where, $f$ is the actual ground truth function, $h_{t+1}$ is the posterior mean of the surrogate at $t+1^{th}$ timestep, $\mathcal{D}_t$ is the training data $\{(x_i, f(x_i))\} \ \forall x \in x_{1:t}$ and $x^*$ is the actual position where $f$ takes the maximum value.

In essence, we are trying to select the point that minimizes the distance to the objective evaluated at the maximum. Unfortunately, we do not know the ground truth function, $f$. ~\cite{mockus1991bayesian} proposed the following acquisition function to overcome the issue.
\begin{equation}
x_{t+1} = argmax_x \mathbb{E} \left( {max} \{ 0, \ h_{t+1}(x) - f(x^+) \} \ | \ \mathcal{D}_t \right)
\end{equation}

Where, $f(x^+)$ is the maximum value that has been encountered so far. This equation for GP surrogate is an analytical expression shown below.

\begin{equation}
EI(x)= \begin{cases} (\mu_t(x) - f(x^+) - \epsilon)\Phi(Z) + \sigma_t(x)\phi(Z), & \text{if}\ \sigma_t(x) > 0 \\ 0, & \text{if}\ \sigma_t(x) = 0 \end{cases}
\end{equation}

\begin{equation}
 Z= \frac{\mu_t(x) - f(x^+) - \epsilon}{\sigma_t(x)}
\end{equation}

where $\Phi(\cdot)$ indicates CDF and  $\phi(\cdot)$ indicates pdf.

From the above expression, we can see that Expected Improvement will be high when: i) the expected value of $\mu_t(x) - f(x^+)$ is high, or, ii) when the uncertainty $\sigma_t(x)$ around a point is high.

Like the PI acquisition function, we can moderate the amount of exploration of the EI acquisition function by modifying $\epsilon$
\subsubsection{\textbf{Upper Confidence Bound (UCB)}}
 One trivial way to come up with acquisition functions is to have a explore/exploit combination.One such trivial acquisition function that combines the exploration/exploitation tradeoff is a linear combination of the mean and uncertainty of our surrogate model. The model mean signifies exploitation (of our model’s knowledge) and model uncertainty signifies exploration (due to our model’s lack of observations).

\begin{equation}
\alpha(x) = \mu(x) + \lambda \times \sigma(x)\alpha(x)=\mu(x)+\lambda×\sigma(x)
\end{equation}

The intuition behind the UCB acquisition function is weighing of the importance between the surrogate’s mean vs. the surrogate’s uncertainty. The $\lambda$ above is the hyperparameter that can control the preference between exploitation or exploration.

We can further form acquisition functions by combining the existing acquisition functions though the physical interpretability of such combinations might not be so straightforward. One reason we might want to combine two methods is to overcome the limitations of the individual methods.
\begin{itemize}
    \item \textbf{Probability of Improvement + $\lambda \ \times$ × Expected Improvement (EI-PI)}:
    
One such combination can be a linear combination of PI and EI. We know PI focuses on the probability of improvement, whereas EI focuses on the expected improvement. Such a combination could help in having a trade-off between the two based on the value of $\lambda$.

\item \textbf{Gaussian Process Upper Confidence Bound (GP-UCB)}: 
Before talking about GP-UCB, let us quickly talk about regret. Imagine if the maximum gold was aa units, and our optimization instead samples a location containing b < ab<a units, then our regret is $a - b$. If we accumulate the regret over $n$ iterations, we get what is called cumulative regret.
GP-UCB’s[11] formulation is given by:
\begin{equation*}
\begin{split}
\alpha_{GP-UCB}(x) = \mu_t(x) + \sqrt{\beta_t}\sigma_t(x)
\end{split}
\end{equation*}
Where $t$ is the timestep.

\end{itemize}

~\cite{srinivas2010gaussian} developed a schedule for $\beta$ that they theoretically demonstrate to minimize cumulative regret.

\subsubsection{\textbf{Entropy search}}
Here, we seek to minimize the uncertainty we have in the
location of the optimal value 
\begin{equation}
    \hat{x} = argmin(f(x))
\end{equation}
Notice that our belief over f induces a distribution over $\hat{x}$, $p(\hat{x}|\mathcal{D}_t)$. Unfortunately, there is no closed-form expression for this distribution.
Entropy search seeks to evaluate points so as to minimize the entropy of the induced distribution
$p(\hat{x}|\mathcal{D}_t)$. Here the utility function is the reduction in this entropy given a new measurement at $x, (x, f(x))$:
\begin{equation}
    u(x) = {H[\hat{x} | \mathcal{D}_t ]}  -  {H[ \hat{x} | \mathcal{D}_t, x, f(x)]}
\end{equation}

As in probability of improvement and expected improvement, we may build an acquisition function by evaluating the expected utility provided by evaluating f at a point x. Due to the nature of the distribution $p(\hat{x}|\mathcal{D}_t)$, this is somewhat complicated, and a series of approximations must be made.

\subsubsection{\textbf{Thompson Sampling}}
Thompson Sampling (TS) \cite{chapelle2011empirical} is a rather old heuristic to address the explore-exploit dilemma that gained popularity in the context of multi-armed bandits. The general idea is to choose an action according to the probability that it is optimal. Applied to BO, this corresponds to sampling a function g from the GP posterior and selecting $x_{t+1} = argmax_{\chi} g$. In fact, let $p_\ast(x)$ be the probability that
point (action) x optimizes the GP posterior, then:


\begin{equation}
\begin{split}
p_*(x) &= \int_{}^{} p_{*}(x|g) p(g|\mathcal{D}_t) dg \\
      &= \int_{}^{} \delta(x - \operatorname*{argmin}_\chi g(x))p(g|\mathcal{D}_t) dg .    
\end{split}
\end{equation}

where $\delta$ denotes the Dirac delta distribution. TS is therefore equivalent to a randomized sequential acquisition function, one that is easy to parallelize. In fact, it suffices to sample multiple functions from the GP posterior to get a batch of points to be evaluated in parallel (P-TS).

\subsubsection{\textbf{Knowledge Gradient}}
Knowledge gradient (KG) ~\cite{frazier2009knowledge} fully accounts for the introduction of noise, and
does not restrict the final recommendation to a previously sampled point. makes
it possible to explore a class of solutions broader than just those that have been previously evaluated when recommending the final solution.
The knowledge gradient policy for discrete $A$ chooses the next sampling
decision by maximizing the expected incremental value of a measurement, without
assuming (as expected improvement does) that the point returned as the optimum
must be a previously sampled point.

\section{Methodology}
\label{Methodology}
Since single acquisition function is not suffice to perform the best over entire optimization, we need to adopt a mixed strategy in which the acquisition function samples from a pool (or portfolio) at each iteration which might work better than any single acquisition. The strategy we need to employ here is very much analogous to a hierarchical multi-armed bandit problem, in which each of the N arms is an acquisition function, each of which is itself an infinite-armed bandit problem. Let us understand the analogy developed for solving bayesian optimization  in the following section. 

\begin{figure*}[t]
	\centering
	\subfloat[Our implementation's results for 'Branin' test function]{%
		\centering \includegraphics[scale=0.5]{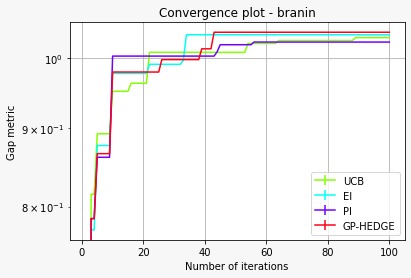} \label{1attrev}}
	\subfloat[Our implementation's Gap-metric results for 'Hartmann6' test function]{%
		\centering \includegraphics[scale=0.5]{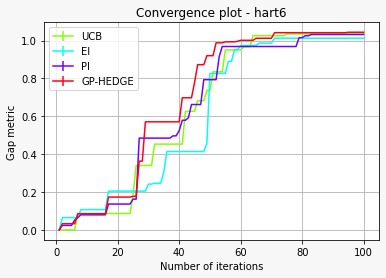} \label{1attadd}}\\
	
	\subfloat[Gap-metric Results from \cite{SirPaper}]{%
		\centering \includegraphics[scale=0.5]{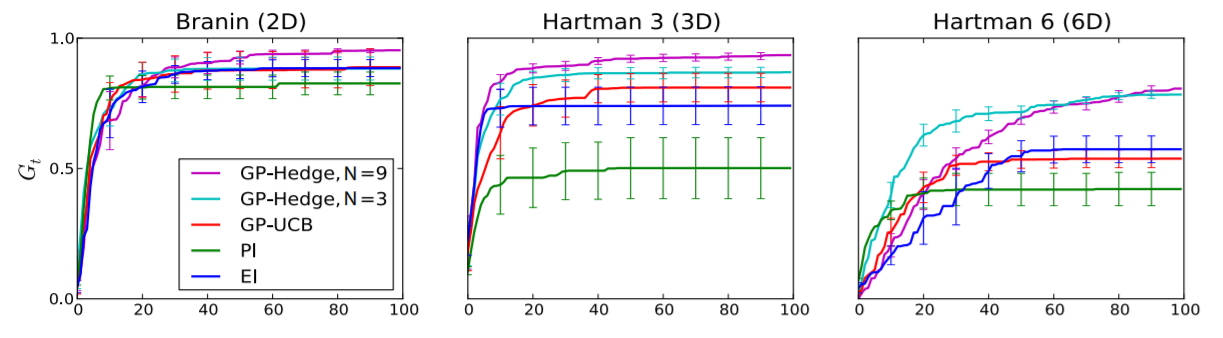} \label{2attrev}}\\
	\subfloat[Cumulative Regret Bound Results from \cite{SirPaper}]{%
		\centering \includegraphics[scale=0.5]{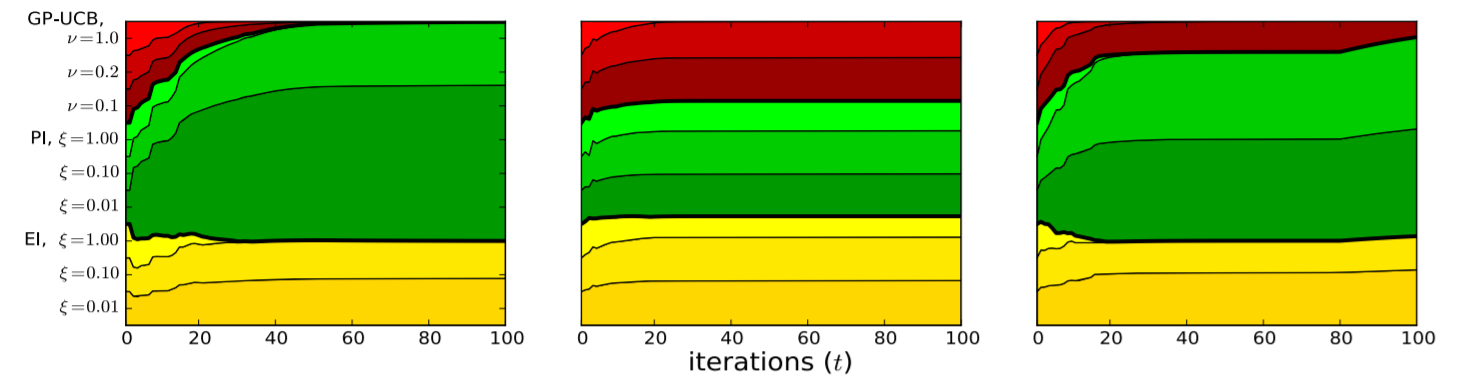} \label{3attrev}}\\
	\caption{Gap-Metric comparison of our implementation and the paper \cite{SirPaper}, and cumulative regret bounds from \cite{SirPaper}}
	\label{exp_results1}
\end{figure*}

\subsection{Portfolio Strategies}
\label{Portfolio Strategies}
The basic idea of portfolio strategy begins with a very popular \emph{hedge} algorithm. In this algorithm, at each time step t  it picks an action i with probability $p_t(i)$ based on the
cumulative rewards (gain) for that action. The term action corresponds to a sample nominated by the following function $u_i$, in this context. Similarly, in bayesian optimization setting, each of N bandits correspond to a single acquisition function.

\begin{equation}
	x_{t}^{i} = argmax_x(u_i(\mathcal{D}_{1:t-1}))
\end{equation}
for  $i=1,2,3....,N$

In GP-hedge after each selection of an action, the algorithm receives reward $r^{i}_{t}$ for each action and updates the gain vector. Thus, \textbf{Hedge} algorithm is based on full information strategy and it requires a reward for every action at every time step. The reward for $i^{th}$ action of $r^{th}$  time step can be denoted by $r_{t}^{i}$ and defined as the expected value of the GP model at $x_{r}^{i}$, i.e $r_{t}^{i} = \mu_{t}(x_{t}^{i})$

A somewhat different algorithm called as \textbf{Exp3}, proposed by ~\cite{portfolio} works on the partial information setting. Unlike GP-hedge the action of each iteration is not rewarded rather only specified actions are rewarded. The algorithm uses Hedge as a
subroutine where rewards observed by Hedge at each iteration are $r^{i}_{t}/\hat{p}_t(i)$ for the action selected and zero for all actions. Here $\hat{p}_t(i)$ is the probability that Hedge would have selected action i. The Exp3 algorithm,
meanwhile, selects actions from a distribution that is
a mixture between $\hat{p}_t(i)$  and the uniform distribution.

\textbf{Algorithm\ref{GPH}} briefs the steps involved in GP-hedge, first of all we select parameter and all gains are initialized to 0. Now for each time step a point has been nominated and nominees are selected with the probability $\hat{p}_t(i)$ as mentioned in line 5 of the algorithm. Sampling of objective function is the next step in this algorithm. Then data augmentation  is further followed by reward distribution and gain updation.

\begin{algorithm}
\SetAlgoLined
\DontPrintSemicolon

Select parameters $\eta \in \mathbb{R}^{+}$ \\
Set $g_0^{i} = 0$ for $i=1, 2..., N$.\\
	\For{$t = 1, 2.... $ \do}{
	Nominate points from each acquisition function:\\
	$X_{t}^{i} =$ argmax$_{x} u_{i} (x|\mathcal{D}_{1:t-1})$\\
	Select nominee $x_t = x_{t}^{j} $ with probability $p_t(j) = \frac{e^{\eta g ^{j}_{t-1}}}{\sum_{l=1}^{k} e^{\eta g^{l}_{t-1}}}$\\
	Sample the objective function $y_t = f(x_t) + \epsilon_t$\\
	Augment the data $\mathcal{D}_{1:t} = \{ \mathcal{D}_{1:t-1}, y_t\}$\\
	Receive rewards $r_{t}^{i} = \mu_t(X_{t}^{i})$ from the updated GP.\\
	Update gains $ g_{t}^{i} = g_{t-1}^{i} + r_{t}^{i}$
	}
\caption{GP-Hedge algorithm}
\label{GPH}
\end{algorithm}

\section{Experimental Results}

\label{Experimental Results}
In this section we describe the experimental settings used in ~\cite{SirPaper} and the experiments replicated by us. For the experimental section we recreate the problem setting as described in ~\cite{SirPaper} and replicate a subset of results provided in the paper. In Section \ref{Experimental Comparison}, we compare the replicated and the original results. In Sections \ref{Test Functions} and \ref{Experimental Settings} we describe the set of test functions used and the environment created.

To test the performance of the optimization method, the method has been tested using a metric known as gap metric ~\cite{EGO-GP}. It is defined as:

\begin{equation}
    G_t = [f(x_+) - f(x_1)] / [f(x_*) - f(X_1)]
\end{equation}

where $x_+$ is the best function sample found at time t, $x_*$ is the known maxima of the test function and $x_1$ is sample at time t = 1.The Gap metric has a range of $[0,1]$ where 0 indicates no improvement over the initial sample and 1 indicates that incumbent is the maximum.

\subsection{Test Functions Used}
\label{Test Functions}
The literature commonly tests Bayesian Optimization on the following three algorithms: Branin, Hartman3, Hartman6. All these functions are continuous, bounded and multimodal with 2,3,6 dimensions respectively. For our replications we use Branin and Hartman6 as test functions. The formulae of the functions can be found here ~\cite{testfunctions}.

\subsection{Experimental Settings}
\label{Experimental Settings}
For the experiments in the paper, number of epochs = 25 and the mean and variance of the gap metric has been computed over time. For GP-Hedge 2 trials have been conducted. In the first trial, 3 acquisition functions have been used Expected Improvement with $\xi = 0.01$, Probability of Improvement with $\xi = 0.01$, Gaussian Process - Upper Confidence Bound (GP-UCB) with $\delta = 0.1$ and $\nu = 0.2$. For the second trial, 9 acquisition functions have been used: Both PI and EI use $\xi = [0.01, 0.1, 1.0]$, GP-UCB uses $\delta = 0.1$ and $\nu = [0.1, 0.2, 1.0]$. In our replication of the results, we omit the 9-acquisition function trial and focus on 3-acquisition function trial. 

\subsection{Experimental Comparison}
\label{Experimental Comparison}
\ref{1attrev} and \ref{1attadd} display the results from the experiments conducted by us \footnote{Code link: \url{https://colab.research.google.com/drive/1GZ14klEDoe3dcBeZKo5l8qqrKf_GmBDn?usp=sharing##scrollTo=XgIBau3O45_V}}. \ref{2attrev} and \ref{3attrev} display the results from the ~\cite{SirPaper} paper. Results from the gap-metric are shown in Fig. \ref{exp_results1}. As we can see from the figure, using a combination of acquisition functions offers the best improvement than using a single acquisition function. This can be seen from the GP-Hedge line in \ref{1attrev} as well as \ref{1attadd} and is corroborated by the results presented in the ~\cite{SirPaper}. These results suggest that the combination of acquisition functions helps in exploring the space early on in the optimization process and slowly as optimization converges focuses on exploitation more. From the cumulative regret bounds we can see that the distribution becomes more stable as the acquisition functions come to a general consensus on the best region to sample. 


\section{Discussions}
\label{Discussions}
This section describes how long does it take for the GP-Hedge Portfolio Allocation Algorithm to reach the optimal solution, i.e., we quantify the convergence of GP-Hedge.

Behaviour of Convergence is difficult to estimate for hedging functions, as past actions effects future state of the problem. Hence, it is not possible to get a link between calculated regret and the regret corresponding to the best acquisition strategy. Also, in order to select an acquisition function, regret bounds are required, which becomes rather difficult due to the difference in the sampled points using GP-Hedge and the the best acquisition strategy solely. However, in order to mitigate these issues, the reward function is approximated as a function of the time-step $t$ is given by the mean as in the following equation -

\begin{equation}
    r_t = \mu_{t-1}(x_t)
\end{equation}

Also, another assumption taken into account is that, GP-UCB is a part of the set of acquisition function., as GP-UCB is shown to have a tight cumulative regret upper bound (as is also evident from its name "Upper Confidence Bound"), which amounts to $\mathcal{O}(\sqrt{T\beta_T\gamma_T})$, $\beta_T$ being the set of hyperparameters of GP-UCB and $\gamma_T$ being the upper bound of the information gain after time-step $T$.

Given all these assumptions, it can be shown that the cumulative regret $R_T$ after time-step $T$ is bounded in the following manner with a very high probability of $(1 - \delta)$, $\delta$ being a small quantity ($\delta > 0$) -

\begin{equation}
    R_T \leq \sqrt{TC_1\beta_T\gamma_T} + [\sum_{t = 1}^{T} \beta_t\sigma_{t-1}(x_t^{UCB})] + \mathcal{O}(\sqrt{T})
\end{equation}

, where $x_t^{UCB}$ is the $t^{th}$ point proposed by GP-UCB.

\section{Future Work}
There is lot of room available for future work in Bayesian optimisation field. Many research directions present themselves in this exciting field. Some of the possibilities are as follows
\begin{enumerate}
    \item Developing Bayesian optimization methods that work well in high dimensions is of great practical and theoretical interest. Directions for research include developing statistical methods that identify and leverage structure present in high-dimensional objectives arising in practice.
    \item It may be particularly fruitful to combine such methodological development with applying Bayesian optimization to important real-world problems, as using methods in the real world tends to reveal unanticipated difficulties and spur creativity.
    \item Substantial impact in a variety of fields seems possible through application of Bayesian optimization. One set of application areas where Bayesian optimization seems particularly well-positioned to offer impact is in chemistry, chemical engineering, materials design, and drug discovery, where practitioners undertake design efforts involving repeated physical experiments consuming years of effort and substantial monetary expense. 
    \item Going beyond Gaussian processes.
\end{enumerate}

\section{Conclusion}
\label{Conclusion}
We present our understanding of Bayesian Optimization in the light of a multi-armed bandit problem. We take the help of the paper - “Portfolio Allocation for Bayesian Optimization” \cite{SirPaper} for the same. We first begin by formalizing the problem of Bayesian Optimization of black-box functions, and the various acquisition functions that are used to act as posterior for a surrogate model used for finding the optimal solution. We also discuss how more than one acquisition function can be selected at a time for sampling using a hierarchical multi-armed bandit framework. Finally, we attempt to reproduce the experiments in the above said paper on known test functions involving optimization using individual acquisition functions, as well as portfolio allocation strategies.

\bibliographystyle{unsrt}  


\end{document}